\documentclass[11pt,a4paper]{article}

\usepackage[hyperref]{acl2021}

\aclfinalcopy

\usepackage{times}
\usepackage{latexsym}

\usepackage[T1]{fontenc}

\usepackage[utf8]{inputenc}

\usepackage{microtype}
\usepackage{subfigure}
\usepackage{hyperref}
\usepackage{graphicx}
\usepackage{multirow}
\usepackage{appendix}
\usepackage{wrapfig,lipsum,booktabs}
\usepackage{color}
\usepackage{url}

\usepackage{soul}
\usepackage{url} 
\usepackage[utf8]{inputenc}
\usepackage{amsmath}
\usepackage{amsthm}
\usepackage{bbm}
\usepackage{commath}
\usepackage{mathtools}

\usepackage{amsfonts}
\usepackage{amssymb}
\usepackage{mathrsfs}
\newtheorem{definition}{Definition}

\usepackage{amsmath,amsfonts,bm}

\def\eqref#1{equation~\ref{#1}}

\def\1{\bm{1}}

\DeclareMathAlphabet{\mathsfit}{\encodingdefault}{\sfdefault}{m}{sl}
\SetMathAlphabet{\mathsfit}{bold}{\encodingdefault}{\sfdefault}{bx}{n}

\title{Temporal-aware Language Representation Learning From Crowdsourced Labels}

 \author{Yang Hao \  \  \ Xiao Zhai  \  \  \  Wenbiao Ding \  \  \  Zitao Liu\thanks{\ \ Corresponding author: Zitao Liu.}  \\
         TAL Education Group, Beijing, China \\
 \texttt{\{haoyang2, zhaixiao, dingwenbiao, liuzitao\}@tal.com}
}

\date{}

\begin{document}
\maketitle
\begin{abstract}

Learning effective language representations from crowdsourced labels is crucial for many real-world machine learning tasks. A challenging aspect of this problem is that the quality of crowdsourced labels suffer high intra- and inter-observer variability. Since the high-capacity deep neural networks can easily memorize all disagreements among crowdsourced labels, directly applying existing supervised language representation learning algorithms may yield suboptimal solutions. In this paper, we propose \emph{TACMA}, a \underline{t}emporal-\underline{a}ware language representation learning heuristic for \underline{c}rowdsourced labels with \underline{m}ultiple \underline{a}nnotators. The proposed approach (1) explicitly models the intra-observer variability with attention mechanism; (2) computes and aggregates per-sample confidence scores from multiple workers to address the inter-observer disagreements. The proposed heuristic is extremely easy to implement in around 5 lines of code. The proposed heuristic is evaluated on four synthetic and four real-world data sets. The results show that our approach outperforms a wide range of state-of-the-art baselines in terms of prediction accuracy and AUC. To encourage the reproducible results, we make our code publicly available at \url{https://github.com/CrowdsourcingMining/TACMA}.

\end{abstract}

\section{Introduction}
\label{sec:intro}
Crowdsourcing offers the ability to utilize the power of human computation to generate data annotations that are needed to train various AI systems. For many practical supervised learning applications, it may be infeasible (or very expensive) to obtain objective and reliable labels due to many reasons such as varying skill-levels and biases of crowdsourced workers. Instead, to improve the quality of labels, we can collect subjective and inconsistent labels from multiple heterogeneous crowdsourced workers. In practice, there is a substantial amount of disagreement between the crowdsourced workers \cite{nie2020can}, i.e., inter-observer variability  or even between a worker and the same worker looking at the same example some time later \cite{guan2018said}, i.e., intra-observer variability. Hence, it is of great practical interest to address supervised learning problems in this scenario. 

Meanwhile, with the recent advances of deep neural networks (DNNs), supervised representation learning (SRL) has led to rapid improvements in the ability of learning intrinsic nonlinear embeddings using DNNs that preserves the distance between similar examples close and dissimilar examples far on the embedding space. In spite of the significant progress for SRL applications such as face recognition \cite{schroff2015facenet}, image retrieval \cite{xia2014supervised}, directly applying existing deep language representation learning approaches on crowdsourced labels may yield poor generalization performance \cite{han2018co}. Because of the high capacity, DNNs could entirely memorize the inconsistency within crowdsourced labels sooner or later during the modeling training process. Besides, this phenomenon does not change with the choice of training optimizations or network architectures  \cite{han2018co}.

A large spectrum of approaches have been successfully developed in either estimating true labels from crowdsourced labels, a.k.a., truth inference or label aggregation \cite{dawid1979maximum,whitehill2009whose}, learning via adversarial data generation \cite{wang2020learning}, or learning language representations discriminatively from large-scale consistent labeled data with complicated neural architectures \cite{rodrigues2018deep}. However, learning effective neural embeddings directly from crowdsourced labels of real-world data poses numerous challenges. First, crowdsourced workers conduct labeling tasks sequentially, i.e., they label samples one after another. Such sequential labeling behavior is a process of learning, and the expertise of the workers is not stable but gradually changing even without feedback \cite{elliott1965effect}. According to Miller's Law \cite{miller1956magical}, humans retain what they just learned in their short-term working memory with a limited span of 7 ± 2. Temporal factors such as fatigue \cite{zhang2018understanding} and intrinsic motivation \cite{kaufmann2011more} implicitly influence the crowdsourcing quality, which are different from existing well-studied factors, such as the quality of crowdsourced workers, the difficulty of data samples, the price of annotation tasks, etc. In the following, such unconscious temporal behaviors are referred to as \textit{``temporal labeling effects''}. How to model such sample-level temporal information for each individual worker undoubtedly poses a hard modeling problem. Second, a large number of real-world crowdsourced data sets have a substantial amount of disagreement among labels and a relatively small sample size. The majority of existing SRL approaches are discriminatively trained on large-scale consistent labeled data to learn their complicated neural architectures, which may easily overfit the inconsistent crowdsourced data. 

In this paper we study and develop solutions that are applicable and can learn effective neural language representations from crowdsourced labels in an end-to-end manner. Our work focuses on the refinements of a popular deep language representation learning paradigm: the deep metric learning (DML) \cite{koch2015siamese,xu2019learning,wang2020representation}. We aim to develop an algorithm to automatically learn a nonlinear language representation of the crowdsourced data from multiple workers using DNNs.

Briefly, the DML is a classical and widely used approach for language representation learning that preserves the distance between similar examples close and dissimilar examples far on the embedding space. The majority of existing DML techniques restricted to just noise-free labels appropriately. However, learning effective representation from highly inconsistent crowdsourced data sets from multiple workers gives rise to numerous important questions: (1) since in practice, annotation performance is affected and varied over time \cite{boksem2005effects,zhang2018understanding}, \emph{how do we capture such temporal labeling effects in the DML learning framework?} (2) while in some cases the problem may be alleviated by pre-processing methods, such as filtering\cite{li2016noise}, label correction\cite{li2019dividemix}, truth inference \cite{dawid1979maximum,raykar2010learning}, etc., the number of remained instances is often significantly reduced or such pre-processing errors for many problems will be propagated to the downstream representation learning tasks. \emph{How to capture the label uncertainties from multiple workers and at the same time prevent the overfitting problem in an end-to-end framework?}

In this work we address the above issues by presenting a \underline{t}emporal-\underline{a}ware language representation learning heuristic for \underline{c}rowdsourced labels with \underline{m}ultiple \underline{a}nnotators (TACMA), that 

\begin{itemize}
	\item utilizes the attention mechanism to capture the temporal influence among sequential labeling tasks according to each worker's short-term working memory.
	\item estimates and aggregates the annotation confidence from disagreements among multiple workers for each sample.
	\item supports language representation learning with DML into an end-to-end fashion, and is extremely easy to implement based on existing DML framework with crowdsourced labels i.e., RLL \cite{xu2019learning}, in around 5 line of codes.
\end{itemize}


\section{Related Work}
\label{sec:related}

\subsection{Truth Inference in Crowdsourcing}
\label{sec:truth}

A large body of research has focused on inferring true labels from crowdsourced labels from multiple workers \cite{dawid1979maximum,whitehill2009whose,li2019exploiting,rodrigues2018deep}. The majority of truth inference approaches are inspired by the classic Expectation-Maximization learning paradigm that iterates between estimating the expertise of annotators given true labels inferred and inferring true labels given the expertise of annotators \cite{dawid1979maximum,whitehill2009whose,zhang2014spectral,li2019exploiting}. Some improvements include modeling the difficulty of items and the expertise of annotators jointly \cite{whitehill2009whose}, applying spectral methods to initialize  worker confusion matrix \cite{zhang2014spectral}, and modeling correlations of workers \cite{li2019exploiting}, etc.

In spite of the successful applications of the truth inference techniques, the majority of aforementioned approaches do not consider the temporal effects of labeling tasks of each individual worker and they cannot seamlessly integrate into deep SRL frameworks.

\subsection{Learning from Noisy Labels}
\label{sec:noisy}

Learning with noisy labels has been an important research topic since the beginning of machine learning \cite{frenay2013classification} and a large spectrum of models have been developed and successfully applied in improving the model prediction performance in noisy settings from different perspectives such as effective label cleaning \cite{lee2018cleannet}, robust model architectures \cite{vahdat2017toward} and loss functions \cite{ghosh2017robust}, sample re-weighting \cite{ren2018learning}, and carefully designed training procedures \cite{zhong2019unequal}. 

However, in this work, different from above approaches of robust learning from noisy labels that assume certain percentage of labels are corrupted, our scenario focuses on noisy labels obtained from multiple annotators where the disagreement (corruption) proportion might be surprisingly high and sometimes even 100\%, i.e., no completely agreement on every single sample from all crowd workers.

\subsection{Deep Metric Learning}
\label{sec:dml}

DML approaches automatically learn nonlinear metric spaces \cite{schroff2015facenet}. Many approaches have achieved promising results in many tasks such as face recognition \cite{schroff2015facenet}, person re-identification \cite{yi2014deep}, and collaborative filtering \cite{hsieh2017collaborative} etc. Recently a body of works have attempted to learn effective embeddings from crowdsourced labels by using DML approaches \cite{xu2019learning,wang2020representation}. For example, Xu et al. estimated crowdsourced label confidence and adjust the DML loss function accordingly \cite{xu2019learning}. An exhaustive review of previous work is beyond the scope of this paper. We refer to the survey of \cite{schroff2015facenet} on works of DML. Although DML approaches are able to learn effective representations, they heavily rely on comparisons within pairs or triplets, which is very sensitive to ambiguous examples and may be easily misled by inconsistent crowdsourced labels.

Please note that models from the above three categories are complementary and they can be combined. For example, learning representation from crowdsourced labels can be conducted in two stages where the truth inference algorithms in Section \ref{sec:truth} is applied to get estimated labels and then the standard DML approaches in Section \ref{sec:dml} are used to output the learned embeddings. Details are discussed in Section \ref{sec:exp}.

\section{The Proposed Approach}
\label{sec:method}
\subsection{Notations}
\label{sec:notation}

Without loss of generality, we consider crowdsourcing scenarios that each data sample is annotated by multiple workers. Following the crowdsourcing practice and to avoid the order effect \cite{hogarth1992order} and cheating, each worker will annotate the same set of samples but with shuffled orders. Let $\boldsymbol{\alpha}^j$ be the sample order index set for the $j^{th}$ worker and $\alpha_i^j$ be the index of $i^{th}$ sample for worker $j$. Let $\mathbf{x}_{\alpha_i^j}$ and $y_{\alpha_i^j}$ be the feature vectors and the worker's assigned label for sample $\alpha_i^j$. Let $\mathcal{F}(\cdot)$ represent the learned language representation. Let $(\cdot)^+$ and $(\cdot)^-$ be the indicators of positive and negative examples. 

\subsection{Temporal-Aware Memory Confidence}
\label{sec:temporal}
According to Miller's Law \cite{miller1956magical}, humans can only hold a very limited number of objects in their short-term working memories. When workers conduct labeling tasks, they tend to make relative comparisons in their memory spans and the annotation quality of one sample is largely influenced by its preceding samples. Therefore, in this work, we focus on studying and modeling the effects of unconscious human behaviors during the labeling process that may implicitly influence the overall crowdsourcing quality. We design an approach to explicitly capture such unconscious temporal human behaviors, i.e., temporal labeling effects. We aim to ensure that \emph{the newly annotated samples should obtain the consistent label with similar samples that have already been annotated recently}. Here we first define the short-term labeling memory as follows:

\begin{definition}{(\textsc{Short-term Labeling Memory})}
\label{def:memory}
A short-term labeling memory of $i^{th}$ sample, i.e., indexed as $\alpha_i^j$, is composed of a sequence of the current item and \textit{k} most recent historical items that have been labeled by worker $j$, i.e., $\mathbf{M}_i^j = \{  <\mathbf{x}_{\alpha_{i}^j}, y_{\alpha_{i}^j}>, <\mathbf{x}_{\alpha_{i-1}^j}, y_{\alpha_{i-1}^j}>, \cdots, <\mathbf{x}_{\alpha_{i-k}^j}, y_{\alpha_{i-k}^j}> \}$. 
\end{definition}

When the new labeling task arrives, i.e., the $i^{th}$ sample, we compute a weight for every element in worker $j$'s short-term labeling memory $\mathbf{M}_i^j$ as the dot product of their learned language representations. This weight might be viewed as an attention over the short-term labeling memory per sample per worker.

To form a proper probability distribution over the elements in $\mathbf{M}_i^j$, we normalize the weights using the softmax function. This way we model probability $s_{\alpha_{i-l}^j}$ that represents the similarity between the $i^{th}$ sample and the sample appears at position $l$ in $\mathbf{M}_i^j$. In a functional form this is:

\vspace{-0.5cm}
\begin{equation}
\nonumber
s_{\alpha_{i-l}^j} \propto \exp \Big( \mathcal{F}(\mathbf{x}_{\alpha_{i-l}^j}), \mathcal{F}(\mathbf{x}_{\alpha_i^j}) \Big), \quad l = 0, \cdots, k
\end{equation}

Then we define a memory confidence score, i.e., $c_{i}^j$, to represent the probability that how likely the sample $i$ is positive ($y_{\alpha_{i}^j} = 1$) solely considering similar samples in the short-term labeling memory. The memory confidence score of $c_{i}^j$ is computed as follows:

\vspace{-1cm}
\begin{equation}
\nonumber
c_{i}^j = \mbox{Pr}(y_{\alpha_{i}^j} = 1) \propto \sum_{l=0}^k \mathbbm{1}\big[y_{\alpha_{i-l}^j} = 1\big] s_{\alpha_{i-l}^j}
\end{equation}

Please note that our attention based temporal-aware memory confidence scores are not limited to binary crowdsourcing tasks and it can be easily extended to multi-class tasks.

\subsection{Multi-Worker Confidence Aggregation}
\label{sec:aggregate}

For each sample $i$, after collecting the memory confidence scores from all workers, we conduct the mean pooling as our aggregation operation, and the final aggregated multi-worker confidence is computed as follows: $c_i = \mbox{MeanPooling}(c_i^1, c_i^2, \cdots, c_i^m)$, where $m$ is the number of workers.

\subsection{Representation Learning Framework}
\label{sec:framework}

We use DML as our representation learning framework. Specifically, following the suggestion of \cite{xu2019learning}, instead of using pair and triplet comparisons, we use group, a.k.a., \emph{n}-tuplet, as our comparison unit. A group is made up of two positive and $n$ negative examples. Similar to \cite{xu2019learning}, we choose to learn our model parameters by maximizing the conditional likelihood of retrieving the positive example $\mathbf{x}_j^+$ given the positive example $\mathbf{x}_i^+$ from a given group. 

Importantly, we do not assume that we know the ground truth label of items in the training set and the validation set. During the training stage of the representation learning framework, after obtaining the aggregated multi-worker confidence $c_i$ of an item with methods introduced in Section \ref{sec:aggregate}, its label is estimated by $\arg\max{c_i}$.

Given a collection of groups, we optimize the DML model parameters by maximizing the sum of log conditional likelihood of finding a positive example $\mathbf{x}_j^+$ given the paired positive example $\mathbf{x}_i^+$ within every group $\mathbf{g}$, which will push items of the same class close and items of different classes far in the embedding space. Furthermore, we incorporate the aggregated temporal-aware multi-worker confidence scores from Section 3.3 into the loss function to capture the inconsistency of crowdsourced labels. The loss function is defined as $\mathcal{L}(\Omega) = - \sum_{}^{} \log p(\mathbf{x}_j^+|\mathbf{x}_i^+)$, 

\vspace{-0.5cm}
\begin{equation}
\label{eq:softmax}
p(\mathbf{x}_j^+|\mathbf{x}_i^+) = \frac{\exp \big(\eta \cdot c_j  \cdot r_{ij}\big)}{\sum_{\mathbf{x}_* \in \mathbf{g}, \mathbf{x}_* \neq \mathbf{x}_i^+} \exp \big(\eta \cdot c_* \cdot r_{i*}\big) } \nonumber
\end{equation}

\noindent where $\Omega$ is the parameter set of the DNN. $r_{i*}$ represents the cosine similarity score between the representations of $\mathbf{x}_i^+$ and $\mathbf{x}_*$  in the embedding space. $\eta$ is a smoothing hyper parameter in the softmax function, which is set empirically on a held-out data set in our experiment. Since $\mathcal{L}(\Omega)$ is differentiable with respect to $\Omega$, we use gradient based optimization approach to train the DNN.

\begin{figure}[!tpbh]
	\centering
	\includegraphics[width=0.46\textwidth] {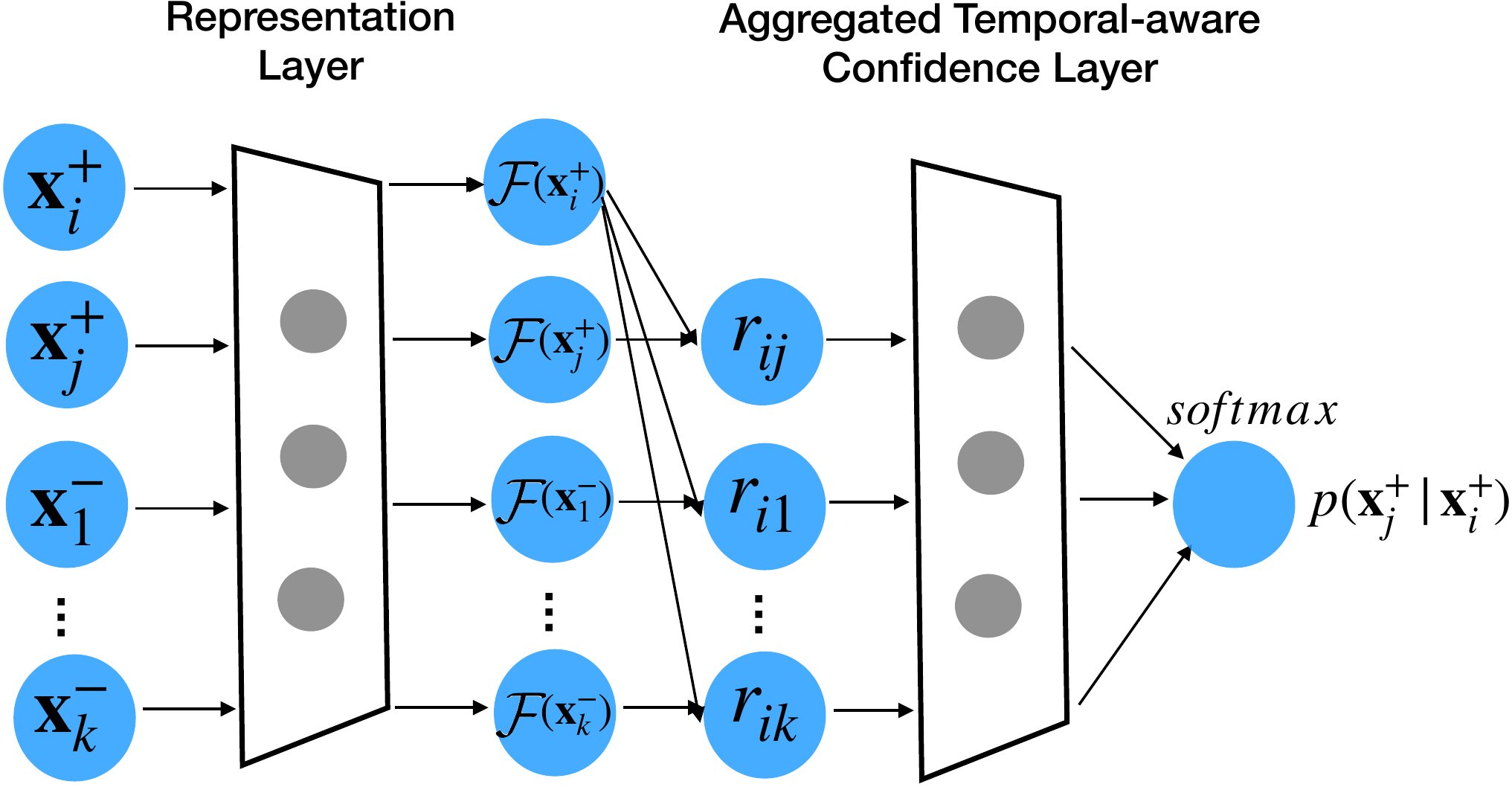}
	\caption{The model structure. Groups made up of two positive and $n$ negative examples are fed into the neural network to obtain their language representations. The cosine similarity scores, i.e., $r_{i*}$, are calculated between the representations of $\mathbf{x}_i^+$ and $\mathbf{x}_*$  in the embedding space. Finally, the goal of training is to maximize the conditional likelihood $p(\mathbf{x}_j^+|\mathbf{x}_i^+)$, which incorporates temporal-aware memory confidence scores $c_{i}^j$.}
	\label{fig:model_structure}
\end{figure}

\section{Experiments}
\label{sec:exp}
Experiments are conducted on both real-world and synthetic data sets. The internal cross validation approach is used to select hyper parameters when optimizing models’ predictive performances. Means as well as standard deviations of both accuracy and AUC scores are reported, to comprehensively evaluate the performance of our proposed method, i.e. \textit{TACMA}. 

\begin{table*}[!thbp]
 	\centering
 	\small
 	\caption{\label{tab:data_stats_horizontal}Data sets statistics.Data sets statistics. It should be noted that the class ratio of each training set is estimated by majority voting since the ground truth labels are not observed. The labels of items in each test set are annotated by experts and they have reached an agreement on the label of each item.}
 	\resizebox{0.97\textwidth}{!}{\begin{tabular}{lllllllll}
 		\toprule
 		
 		\textbf{Data Sets}         & \textbf{Emotion} & \textbf{Commending} & \textbf{Question} & \textbf{Concluding}& \textbf{Syn-2}   & \textbf{Syn-4}      & \textbf{Syn-6}    & \textbf{Syn-8}  \\
 		\midrule
 		\# of annotators         & 5                & 7                   & 5                 & 5                & 7                & 9                   & 11                & 13    \\
 		\# of train samples      & 3067             & 1200                & 3140              & 1208         & 3140             & 3140                & 3140              & 3140                     \\
 		\# of validation samples & 766              & 299                 & 785               & 302     & 785              & 785                 & 785               & 785                           \\
 		\# of test samples       & 800              & 1300                & 2000              & 648           & 2000             & 2000                & 2000              & 2000             \\
 		kappa                    & 0.84             & 0.69                & 0.82              & 0.37   & 0.35             & 0.2                 & 0.12              & 0.08                  \\
 		train class ratio (majority voting)             & 0.42         & 0.50            & 0.63          & 0.42       & 0.63             & 0.63                & 0.63              & 0.63        \\
 		
 		\bottomrule
 	\end{tabular}}
 \end{table*}

\subsection{Real-World Data Sets}
Experiments are first conducted on 4 real-world data sets and the corresponding descriptive statistics can be found in Table \ref{tab:data_stats_horizontal}. 

\begin{itemize}
	\item \textit{Emotion}: A vocal emotional speech data set with binary labels indicating whether the voice fragment is exciting or not.
	\item \textit{Concluding}: A linguistic data set where each item is labeled on whether it is a conclusion of a lesson.
	\item \textit{Commending}: A linguistic data set of ASR transcripts from real-world classroom recordings. Each item is labeled on whether it's a commending instruction from the  instructors. 
	\item \textit{Question}: A vocal speech data set where each item is labeled on whether it is an interrogative sentence.
\end{itemize}

Acoustic features of the \textit{Emotion} data set are extracted using OpenSmile\footnote{https://www.audeering.com/opensmile/} with the computational paralinguistic challenge's (COMPARE-2013) feature set	\cite{schuller2013interspeech}. Sentence embedding features are extracted with a Chinese RoBERTa pretrained model \footnote{https://github.com/ymcui/Chinese-BERT-wwm}. Again we emphasize that the ground truth labels of items in the training and validation set are not observed. In order to evaluate the performance of each model objectively, the labels of items in test sets are labeled by experts and they have reached an agreement on the labels of items.

Inter-observer variability of each data set is measured with Fleiss-kappa score \cite{fleiss1971measuring}. Intra-observer variability, i.e., the level of consistency of an annotator when labeling items from the same class, is hard to directly measure without ground truth labels. We will explore the effect of intra-observer variability using temporal-aware memory confidence in Section \ref{sec:corr_acc_confidence}. 

\subsection{Synthetic Data Sets}
In real-world scenarios, annotators are not guaranteed to be serious about their annotating work, and one may assign random labels in order to get paid quickly. Methods designed for crowdsourcing scenarios should be able to get rid of the influence of these noisy annotations. Hence we build synthetic data sets to evaluate the robustness to irresponsible annotators of each method. Starting from the original \textit{Question} data set, we gradually add 2, 4, 6 and 8 simulated irresponsible annotators. They make random judgments regardless of the features of items. Hence in the worst case, 8 out of 13 workers are making random judgments, resulting in an extreme low kappa of 0.02. Experiments conducted on these synthetic data sets are helpful to examine the robustness of methods.

\subsection{Baselines}
We carefully selected several groups of baselines as follows:

\textbf{Group 1: Truth Inference.} A wide range of label aggregation methods are chosen as our baselines. Some widely-used methods according to the survey \cite{zheng2017truth} are included, i.e., \emph{EM} \cite{dawid1979maximum}, \emph{Spectral-EM}\cite{zhang2014spectral}, \emph{GLAD} \cite{whitehill2009whose}, \emph{IBCC} \cite{kim2012bayesian}, \emph{VI-BP} \cite{qiang2012variational}, \emph{VI-MF} \cite{qiang2012variational}, \emph{KOS} \cite{david2011iterative}, \emph{ZenCrowd} \cite{demartini2012zencrowd}, \emph{LFC} \cite{raykar2010learning}, \emph{PM} \cite{li2014resolving}, and the implementation of these algorithms can mostly be found in the website\footnote{https://zhydhkcws.github.io/crowd\_truth\_inference/index.html}. Meanwhile some more recent works are also included: \emph{EBCC} \cite{li2019exploiting}, \emph{BWA} \cite{li2019truth}. 

\textbf{Group 2: Representation Learning.} Our proposed method is compared with representation learning methods via deep metric learning, including Triplet with semi-hard example mining \cite{schroff2015facenet}, i.e., \emph{Triple}, and Triplet networks with Center Loss \cite{he2018triplet}, i.e., \emph{Center}. Recent works of learning effective embeddings from crowdsourced labels using DML approaches are also important baselines: RLL-MLE \cite{xu2019learning}, RLL-Bayesian \cite{xu2019learning},  RECLE \cite{wang2020representation}. 

\textbf{Group 3: Learning from Noisy Data.} Group 3 contains methods of learning with noisy labels: \emph{LC} \cite{arazo2019unsupervised} use a two-component beta mixture model to perform unsupervised noise modeling, and \emph{DivideMix} \cite{li2019dividemix} leverages semi-supervised learning techniques. \emph{CrowdLayer} \cite{rodrigues2018deep} is an end-to-end approach learning a DNN from noisy labels with a crowd layer.

\textbf{Group 4: Combining Group 1 with Groups 2 \& 3.} Some methods of Group 2 \& 3, i.e. \emph{Triple}, \emph{Center}, \emph{LC}, \emph{DivideMix}, are not specifically designed for crowdsourcing scenarios. Although majority-voting labels are served as a default choice, these models should be trained with labels inferred by methods of Group 1 as stronger baselines, since methods of Group 1 are likely to provide more accurate inferred labels than majority voting. These methods are therefore trained with labels inferred by \emph {EBCC}, which achieves the best performances of Group 1 in all data sets.

\subsection{Setup and Implementation Details}
\label{sec:implementation}


Experimental codes are implemented in Tensorflow 1.8 available at \url{https://github.com/CrowdsourcingMining/TACMA}. Experiments are conducted on a server with a GTX 1080 Ti GPU. We set the tuplet size $n$ to 5 for all the experiments, as suggested in \cite{xu2019learning}. The representation learning network has a simple structure, i.e., 2 fully-connected layers with a drop-out rate of 0.2, a learning rate of 1e-3, and hyper-parameters including sizes of each layer and scale of $\ell_2$ regularization searched via grid searching with cross validation. The network weights are initialized with a normal distribution initializer and updated with Adadelta optimizer \cite{zeiler2012adadelta}. For all the representation learning methods, the downstream classifier is set to be a logistic regression classifier with $\ell_2$ penalty containing the only hyper-parameter $C$ as penalty strength ranging from 1e-2 to 1e4.

\subsection{Performance Comparison}

\begin{table*}[t]
\setlength \tabcolsep{3pt}
	\centering
\small
\caption{\label{tab:performance}Prediction accuracy and AUC scores on 4 real-world data sets. The experiments are repeated 5 times and the means and standard deviations are reported.}
\resizebox{0.97\textwidth}{!}{\begin{tabular}{lllllllll}
				\toprule
                & \multicolumn{2}{c}{Commending}              & \multicolumn{2}{c}{Emotion}                 & \multicolumn{2}{c}{Question}                & \multicolumn{2}{c}{Concluding} \\
                \cmidrule(lr){2-3} \cmidrule(lr){4-5} \cmidrule(lr){6-7} \cmidrule(lr){8-9}
                & \multicolumn{1}{c}{ACC} & \multicolumn{1}{c}{AUC} & \multicolumn{1}{c}{ACC} & \multicolumn{1}{c}{AUC} & \multicolumn{1}{c}{ACC} & \multicolumn{1}{c}{AUC} & \multicolumn{1}{c}{ACC} & \multicolumn{1}{c}{AUC} \\
				\midrule 
EM                    & 0.794±0.019          & 0.871±0.008          & 0.883±0.012          & 0.967±0.005          & 0.877±0.010          & 0.941±0.005          & 0.681±0.004          & 0.720±0.015          \\
Spectral-EM           & 0.794±0.017          & 0.870±0.007          & 0.886±0.010          & 0.964±0.003          & 0.876±0.009          & 0.941±0.004          & 0.681±0.004          & 0.720±0.013          \\
GLAD                  & 0.794±0.017          & 0.870±0.007          & 0.886±0.010          & 0.964±0.003          & 0.878±0.009          & 0.942±0.004          & 0.689±0.009          & 0.742±0.014          \\
IBCC                  & 0.794±0.017          & 0.870±0.007          & 0.889±0.004          & 0.964±0.004          & 0.876±0.009          & 0.941±0.004          & 0.681±0.004          & 0.720±0.013          \\
VI-BP                 & 0.794±0.017          & 0.870±0.007          & 0.892±0.012          & 0.968±0.005          & 0.877±0.008          & 0.941±0.004          & 0.681±0.004          & 0.720±0.013          \\
VI-MF                 & 0.799±0.013          & 0.874±0.003          & 0.786±0.000          & 0.898±0.000          & 0.876±0.009          & 0.941±0.004          & 0.685±0.003          & 0.725±0.009          \\
KOS                   & 0.799±0.013          & 0.874±0.003          & 0.786±0.000          & 0.898±0.000          & 0.878±0.009          & 0.942±0.004          & 0.694±0.005          & 0.747±0.010          \\
ZenCrowd              & 0.794±0.019          & 0.871±0.008          & 0.895±0.011          & 0.971±0.004          & 0.877±0.010          & 0.941±0.005          & 0.689±0.010          & 0.742±0.016          \\
LFC                   & 0.794±0.019          & 0.871±0.008          & 0.883±0.009          & 0.967±0.004          & 0.877±0.010          & 0.941±0.005          & 0.681±0.004          & 0.720±0.015          \\
PM                    & 0.799±0.014          & 0.867±0.008          & 0.887±0.010          & 0.966±0.003          & 0.874±0.010          & 0.940±0.004          & 0.677±0.009          & 0.730±0.013          \\
EBCC                  & 0.812±0.006          & 0.874±0.003          & 0.895±0.012          & 0.970±0.005          & 0.878±0.007          & 0.941±0.008          & 0.694±0.003          & 0.748±0.006          \\
BWA                   & 0.794±0.020          & 0.867±0.008          & 0.888±0.005          & 0.965±0.004          & 0.875±0.009          & 0.939±0.004          & 0.689±0.007          & 0.741±0.013          \\
\midrule
Triple                & 0.793±0.012          & 0.871±0.006          & 0.804±0.005          & 0.876±0.002          & 0.888±0.002          & 0.941±0.001          & 0.725±0.014          & 0.821±0.008          \\
Center       & 0.806±0.002          & 0.859±0.001          & 0.701±0.007          & 0.780±0.007          & 0.840±0.006          & 0.905±0.008          & 0.705±0.015          & 0.797±0.007          \\
RLL-MLE               & 0.800±0.008          & 0.866±0.001          & 0.854±0.016          & 0.961±0.008          & 0.853±0.013          & 0.919±0.006          & 0.735±0.004          & 0.828±0.009          \\
RLL-Bayesian          & 0.816±0.000          & 0.861±0.001          & 0.877±0.006          & 0.954±0.004          & 0.877±0.004          & 0.932±0.003          & 0.725±0.001          & 0.839±0.001          \\
RECLE                 & 0.812±0.002          & 0.858±0.000          & 0.746±0.001          & 0.836±0.001          & 0.880±0.024          & 0.934±0.012          & 0.729±0.003          & 0.838±0.005          \\
\midrule
LC                    & 0.560±0.085          & 0.700±0.028          & 0.611±0.046          & 0.715±0.007          & 0.715±0.018          & 0.720±0.007          & 0.701±0.018          & 0.790±0.011          \\
DivideMix             & 0.515±0.016          & 0.733±0.014          & 0.535±0.000          & 0.730±0.000          & 0.734±0.009          & 0.720±0.014          & 0.654±0.025          & 0.710±0.007          \\
CrowdLayer            & 0.802±0.008          & 0.878±0.007          & 0.757±0.008          & 0.798±0.009          & 0.852±0.003          & 0.920±0.002          & 0.676±0.014          & 0.722±0.011          \\
\midrule
LC+EBCC               & 0.581±0.070          & 0.687±0.029          & 0.825±0.024          & 0.845±0.019          & 0.758±0.010          & 0.830±0.012          & 0.705±0.018          & 0.784±0.004          \\
DivideMix+EBCC        & 0.515±0.018          & 0.730±0.014          & 0.726±0.039          & 0.832±0.028          & 0.760±0.012          & 0.833±0.014          & 0.659±0.006          & 0.720±0.005          \\
Triple+EBCC           & 0.814±0.004          & 0.872±0.000          & 0.893±0.003          & 0.968±0.004          & 0.890±0.001          & 0.938±0.003          & 0.737±0.003          & 0.825±0.007          \\
Center+EBCC  & 0.814±0.004          & 0.866±0.003          & 0.826±0.016          & 0.884±0.018          & 0.844±0.005          & 0.909±0.005          & 0.742±0.006          & 0.848±0.003          \\
\midrule
TACMA            & \textbf{0.831±0.002} & \textbf{0.882±0.004} & \textbf{0.904±0.002} & \textbf{0.973±0.001} & \textbf{0.899±0.005} & \textbf{0.945±0.003} & \textbf{0.765±0.006} & \textbf{0.855±0.010}\\
\bottomrule
\end{tabular}}
\end{table*}

\begin{figure*}[t]\centering                                                          
\subfigure[Raw Embedding]{                    
	\begin{minipage}{2.8cm}\centering \includegraphics[width=1.1\textwidth]{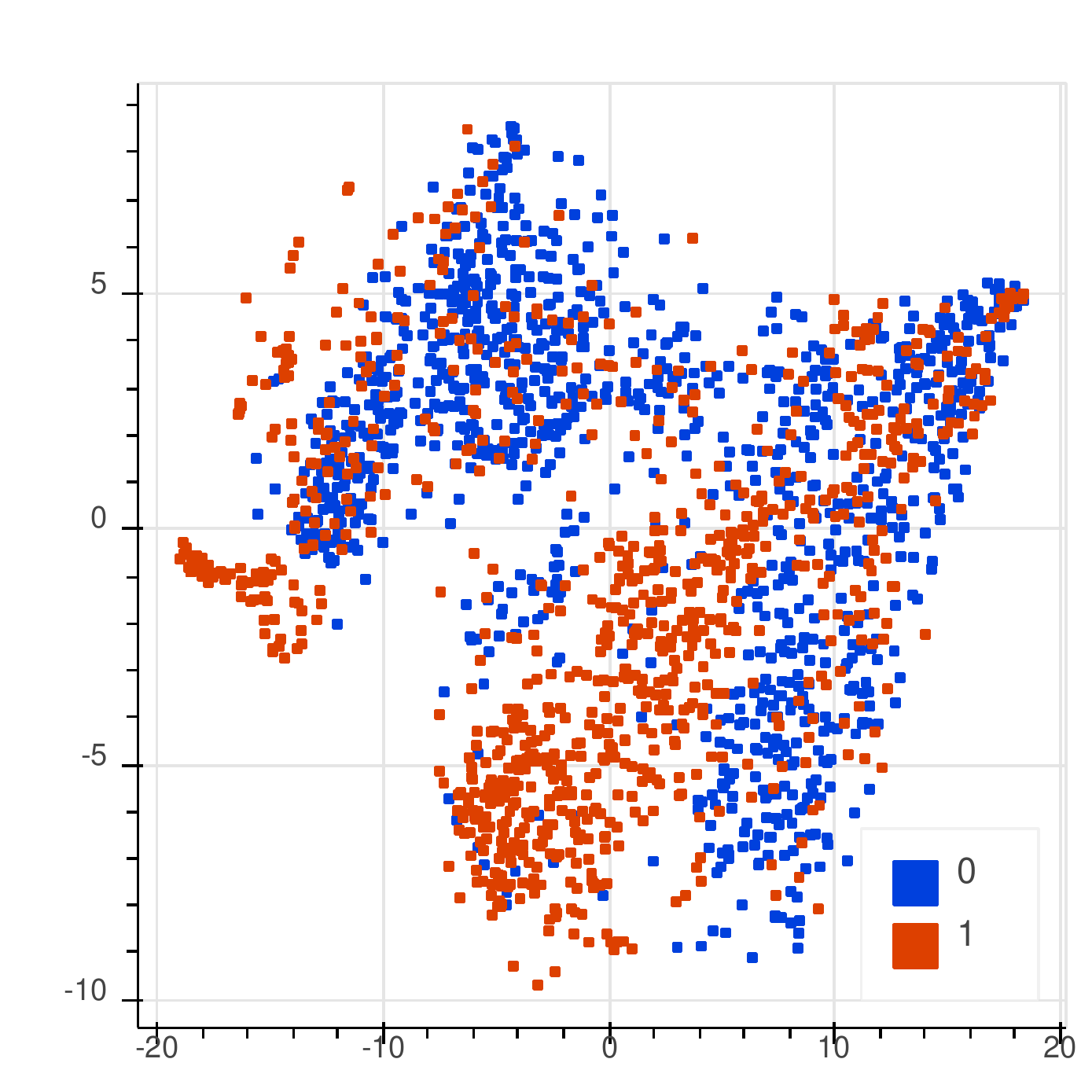}
   \end{minipage}}
\subfigure[RLL-MLE]{                   
 \begin{minipage}{2.8cm}\centering  \includegraphics[width=1.1\textwidth]{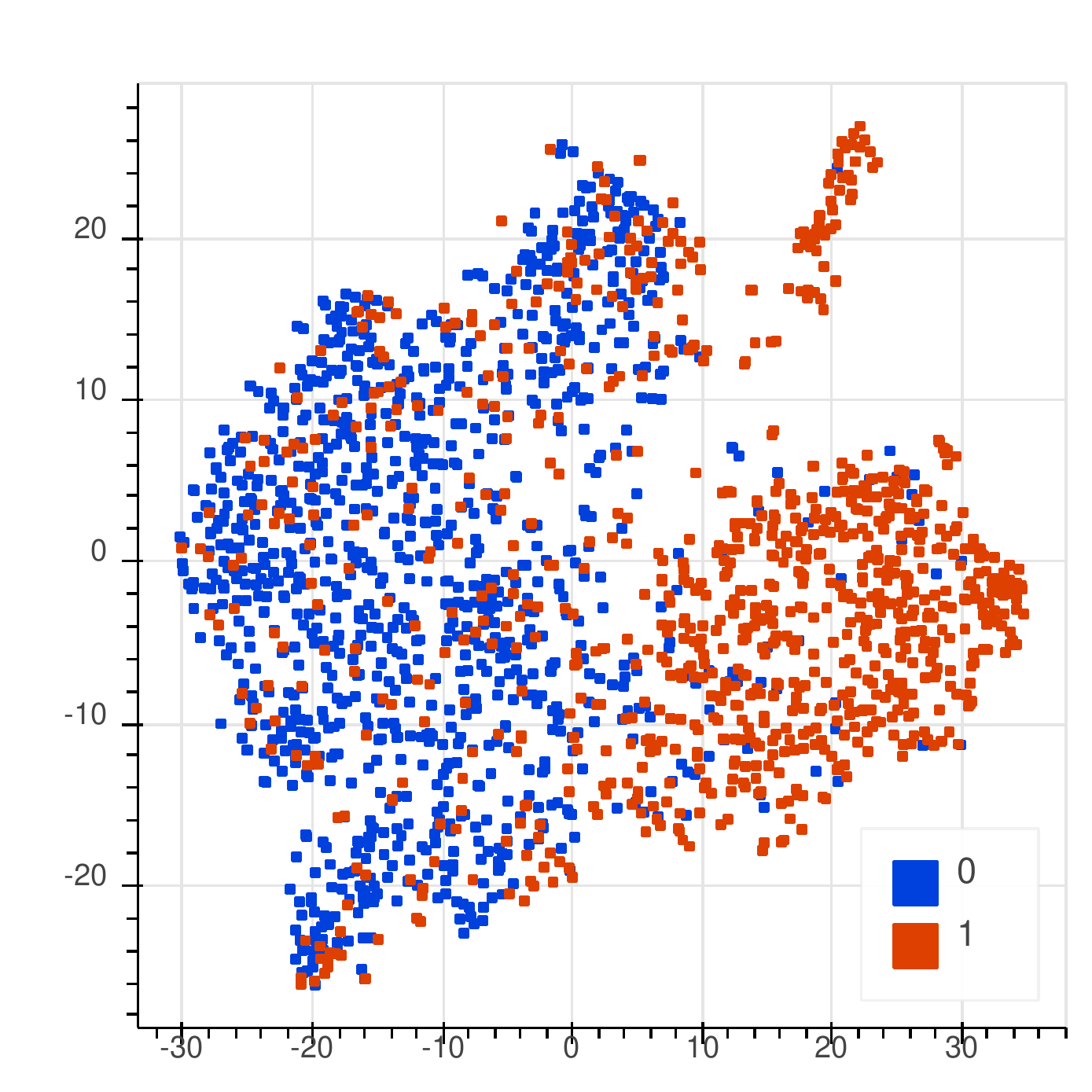}                
\end{minipage}}
\subfigure[RLL-Bayesian]{                   
	\begin{minipage}{2.8cm}\centering  \includegraphics[width=1.1\textwidth]{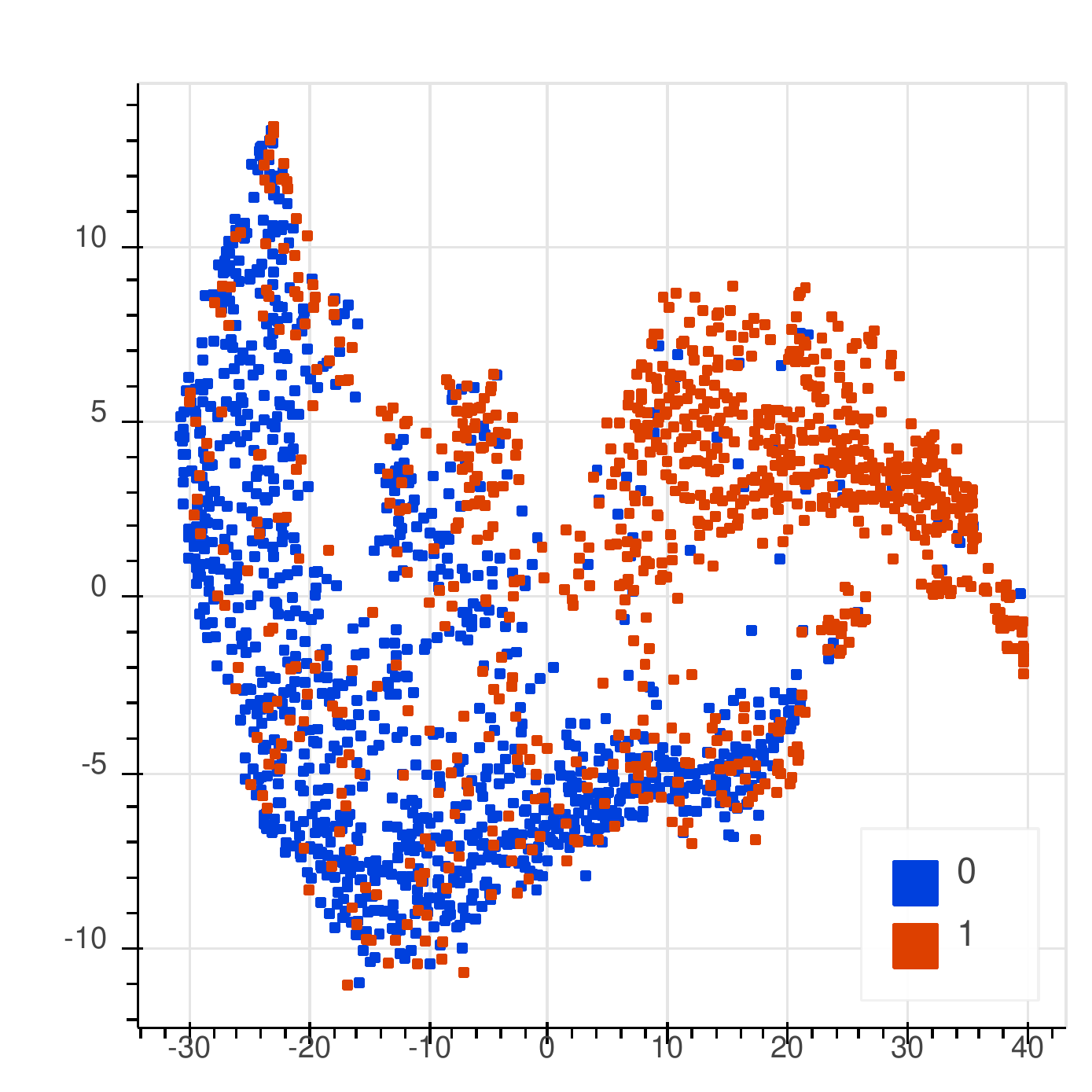}                
\end{minipage}}
\subfigure[RECLE]{
	\begin{minipage}{2.8cm}\centering  \includegraphics[width=1.1\textwidth]{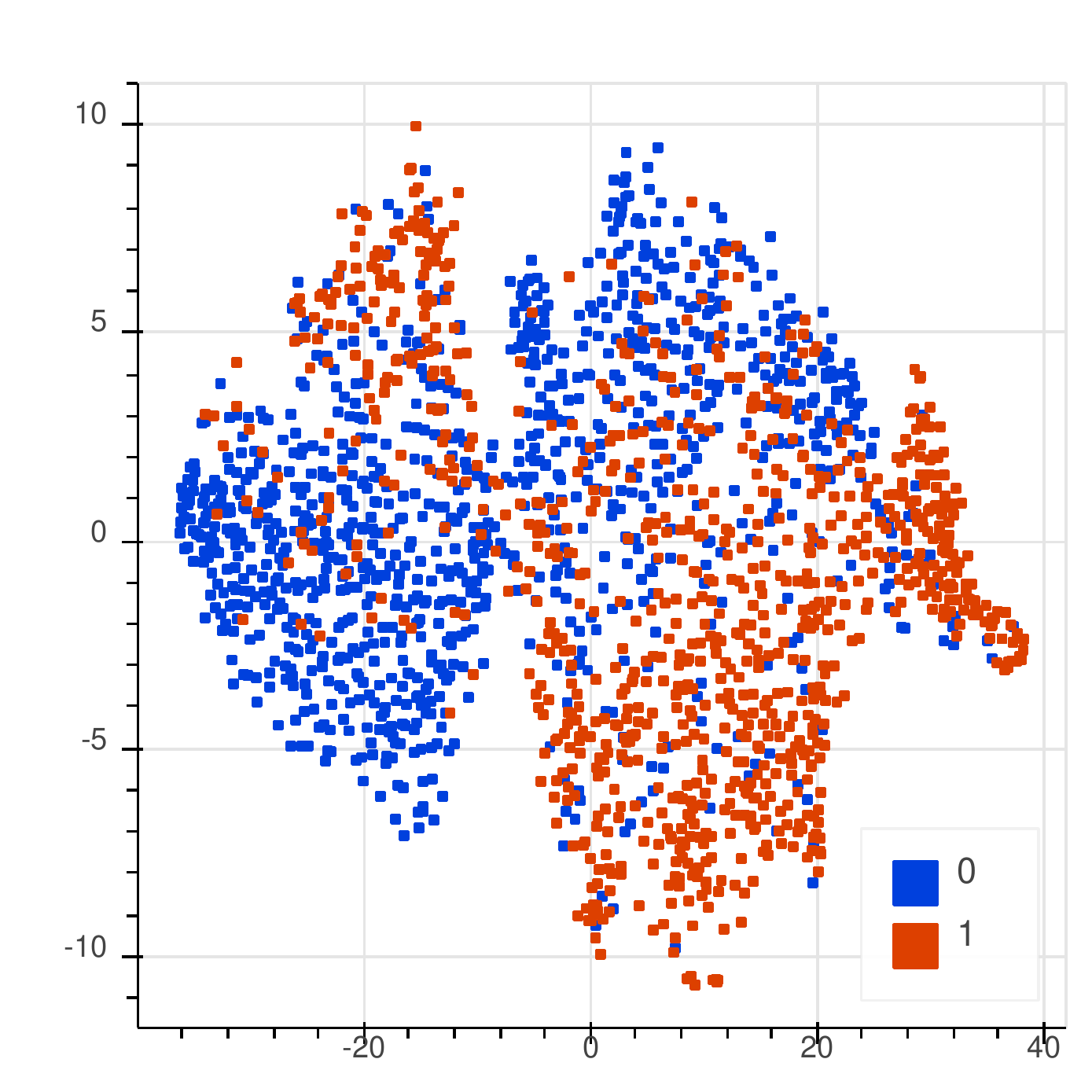}                
\end{minipage}}
\subfigure[TACMA]{
	\begin{minipage}{2.8cm}\centering  \includegraphics[width=1.1\textwidth]{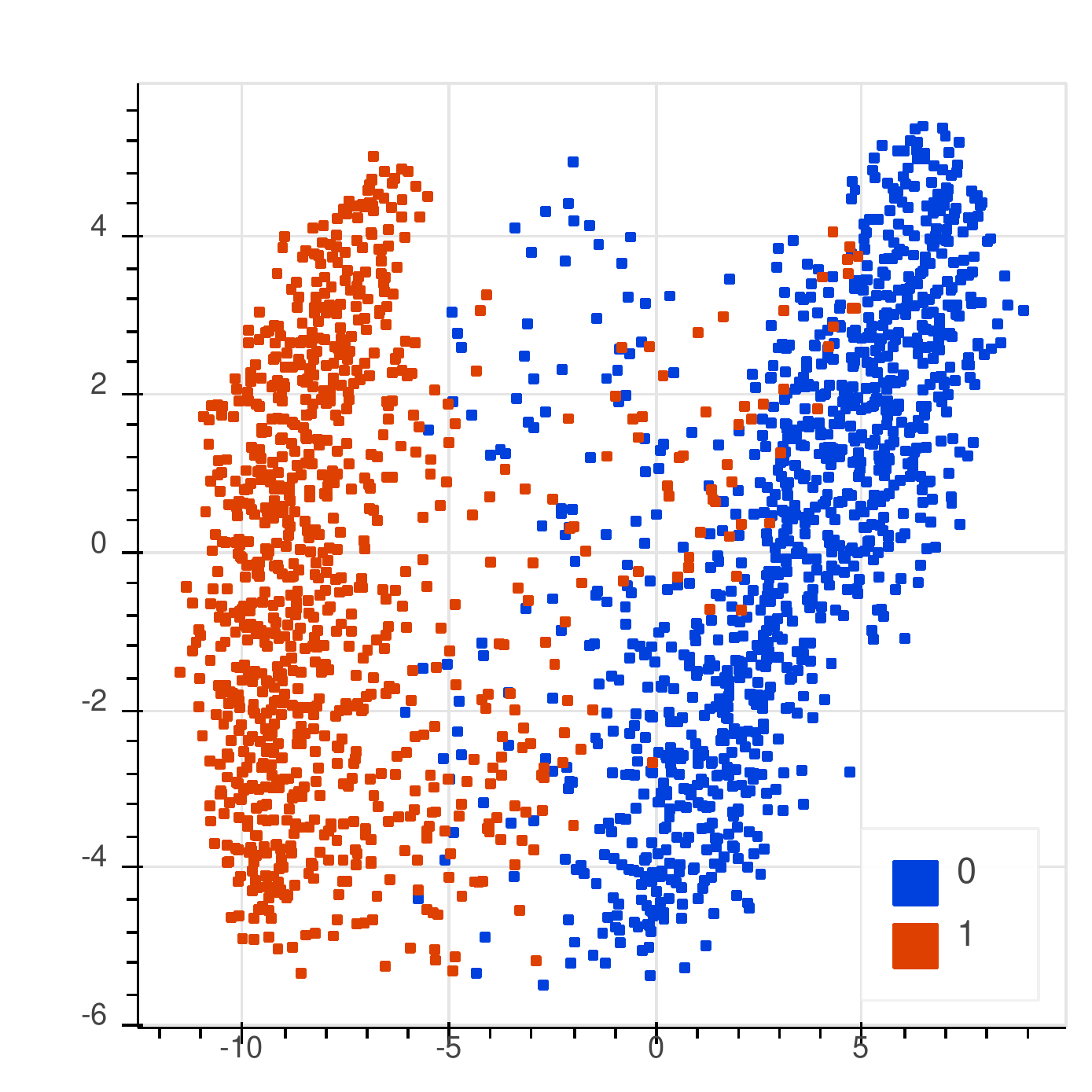}   
\end{minipage}}
\caption{Visualization of learned representations on the test set of Question data. The raw features  of items are fed into the representation network to obtain the semantic representations, and  dimension reduction using t-SNE method is performed for visualization. }
	\label{fig:representation}
	\vspace{-0.8cm}
\end{figure*}

We compare performance of \textit{TACMA} with existing methods on 4 real-world data sets and the results are summarized in Table \ref{tab:performance}. \textit{TACMA} outperforms all the 4 groups of baselines, and here are some observations:

\begin{itemize}
	\item The advantage of \textit{TACMA} over truth inference methods gets bigger on the \textit{Concluding} data set than other data sets. The \textit{Concluding} data set has a low kappa score of 0.37, indicating that there are more disagreements among workers, which makes it hard to inference correct labels regardless of items' features. By contrast, \textit{TACMA} makes full use of representations of items to gain more information resulting in the best performance.
	
	\item Although labels inferred by \textit{EBCC} boost the performances of representation learning models, e.g., \textit{Triple+EBCC}, they still perform inferior to \textit{TACMA}, a possible explanation is that these two-stage methods give equal weight to each item and ignores temporal labeling effects. \textit{TACMA} is able to discover potential conflicts in the short-term working memory, by applying the attention mechanism and gives low weights to the conflicting judgments.
	
	\item \textit{TACMA} shares the same representation network structures with other methods of representation learning with crowdsourced labels i.e., \textit{RLL-MLE}, \textit{RLL-bayesian} and \textit{RECLE}. The learned representations are compared in Figure \ref{fig:representation} by feeding the raw features into representation network and performing dimension reduction into 2-dimensional space with t-SNE method \cite{van2008visualizing}. In the raw feature space, items of different classes are interleaved with each other. By contrast, learned representations of \textit{TACMA} are more separated than the other methods, reducing the difficulty of downstream classification tasks.
\end{itemize}

\begin{figure}[!tpbh]
	\centering
	\includegraphics[width=0.5\textwidth] {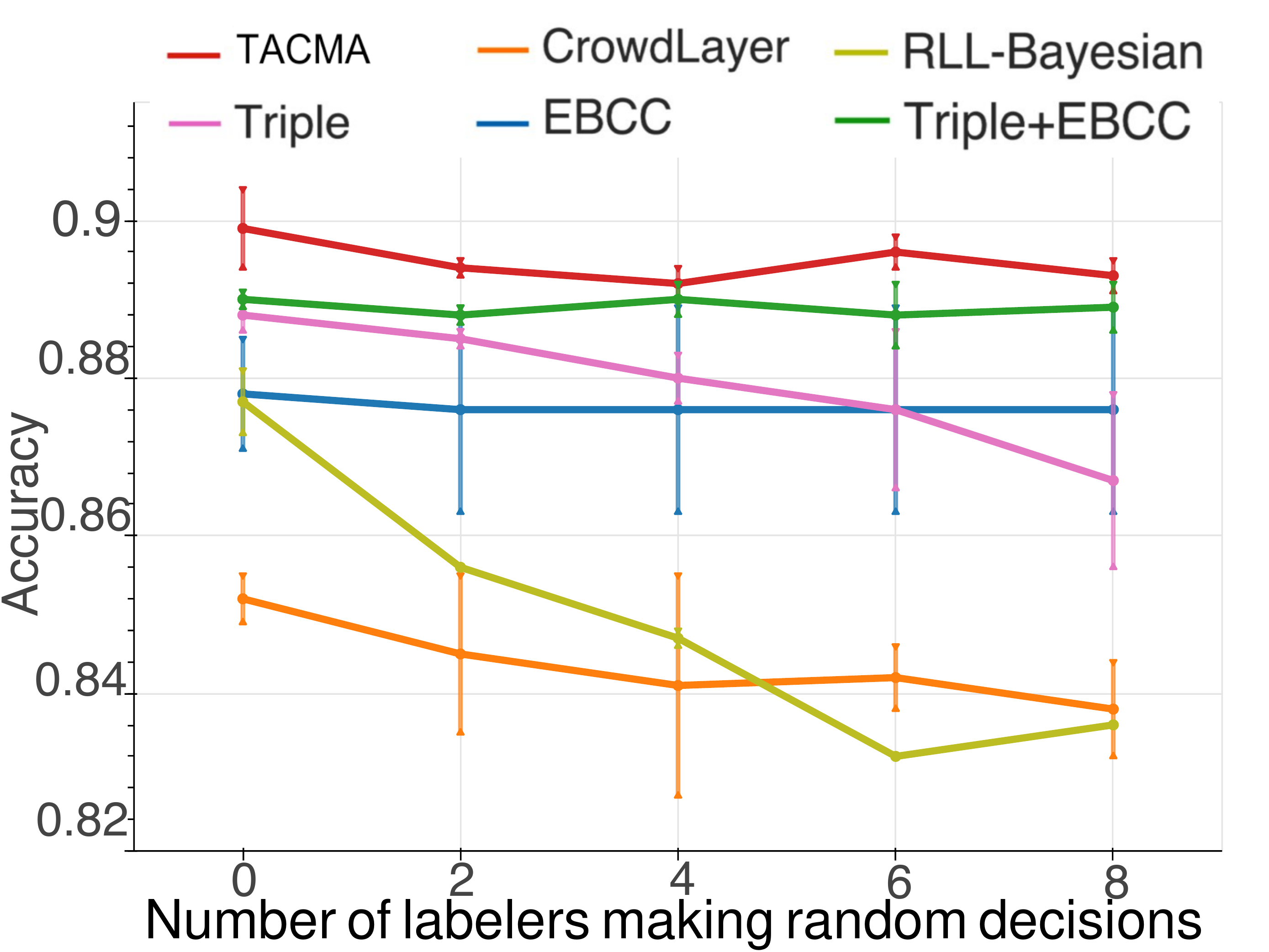}
	\caption{Accuracy curves on synthetic data sets containing different number of irresponsible annotators who make random decisions.}
	\label{fig:add_randoml}
\end{figure}

\subsection{Robustness to Irresponsible Workers}

We select some representatives from Groups 1-4 and draw the curves of accuracy on synthetic data sets containing different number of irresponsible workers in Figure \ref{fig:add_randoml}. We can find that:

\begin{itemize}
	\item Truth inference methods such as \textit{EBCC} stay stable facing different numbers of irresponsible workers. On the other hand, the accuracy of other methods decreases when increasing the number of irresponsible workers. This result may be explained by the fact that for methods including \textit{RLL-Bayesian}, \textit{Triple}, learning effective representations of items heavily relies on correct labels,	and hence becomes harder as the labels become more noisy.
	
	\item \textit{TACMA} maintains the highest accuracy of all the methods. Unlike the two-stage method i.e., \textit{Triplet + EBCC}, which gives equal weight to each item and ignores \textit{temporal labeling effects},  \textit{TACMA} is able to discover potential conflicts in the short-term working memory using the attention mechanism, and give low training weights to the conflicting judgments. 
\end{itemize}

\subsection{Effect of Working Memory Sizes}
\begin{figure}[!tpbh]
	\centering
	\includegraphics[width=0.47\textwidth] {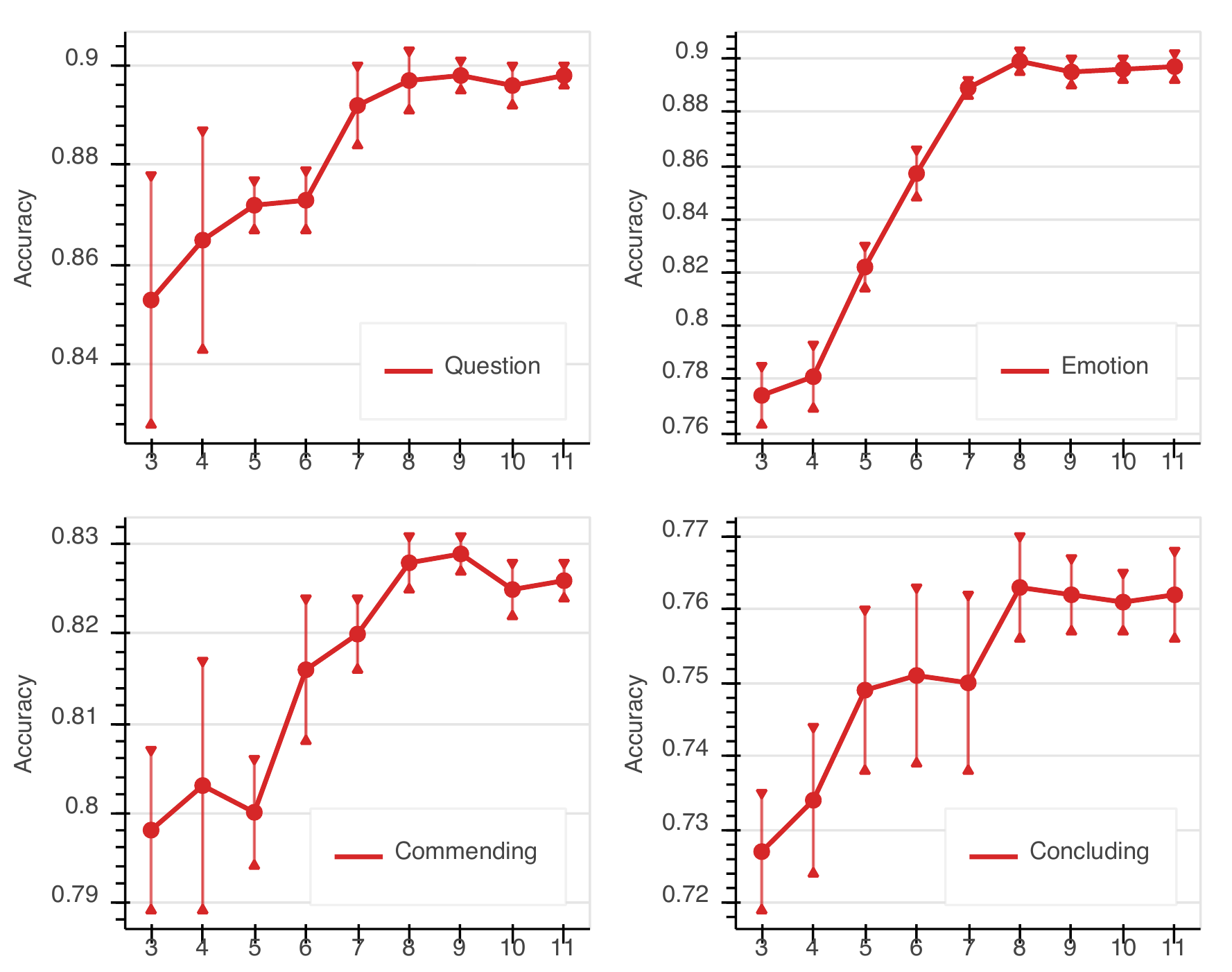}
	\caption{The effect of different working memory sizes on prediction accuracy on real-world data sets.}
	\label{fig:window_size_4_datasets}
\end{figure}

We set the working memory size ranging from 3 to 11 to find the optimized length and at the same time explore its influence on performance, shown in Figure \ref{fig:window_size_4_datasets}. The accuracy of our proposed method goes up at the beginning with the increasing working memory size, and the standard deviations gradually become smaller at the same time. It is reasonable because potential inconsistent judgments among similar items cannot be found without observing enough historical annotations. As the working memory size continues extending, the accuracy scores become relatively stable, indicating that there is sufficient evidence to estimate the time-aware confidence of the current annotation.

\subsection{Relations between Temporal-aware Memory Confidence and Worker's Expertise}
\label{sec:corr_acc_confidence}

In this part we further explore the relations between worker's expertise and temporal-aware memory confidence. To evaluate a worker's expertise, a Logistic Regression classifier is trained with labels annotated by this same person, and the accuracy on the corresponding test set is recorded. On the other hand, the temporal-aware confidence of all the judgments made by this worker is averaged. 

\begin{figure}[!tpbh]
	\centering
	\includegraphics[width=0.38\textwidth] {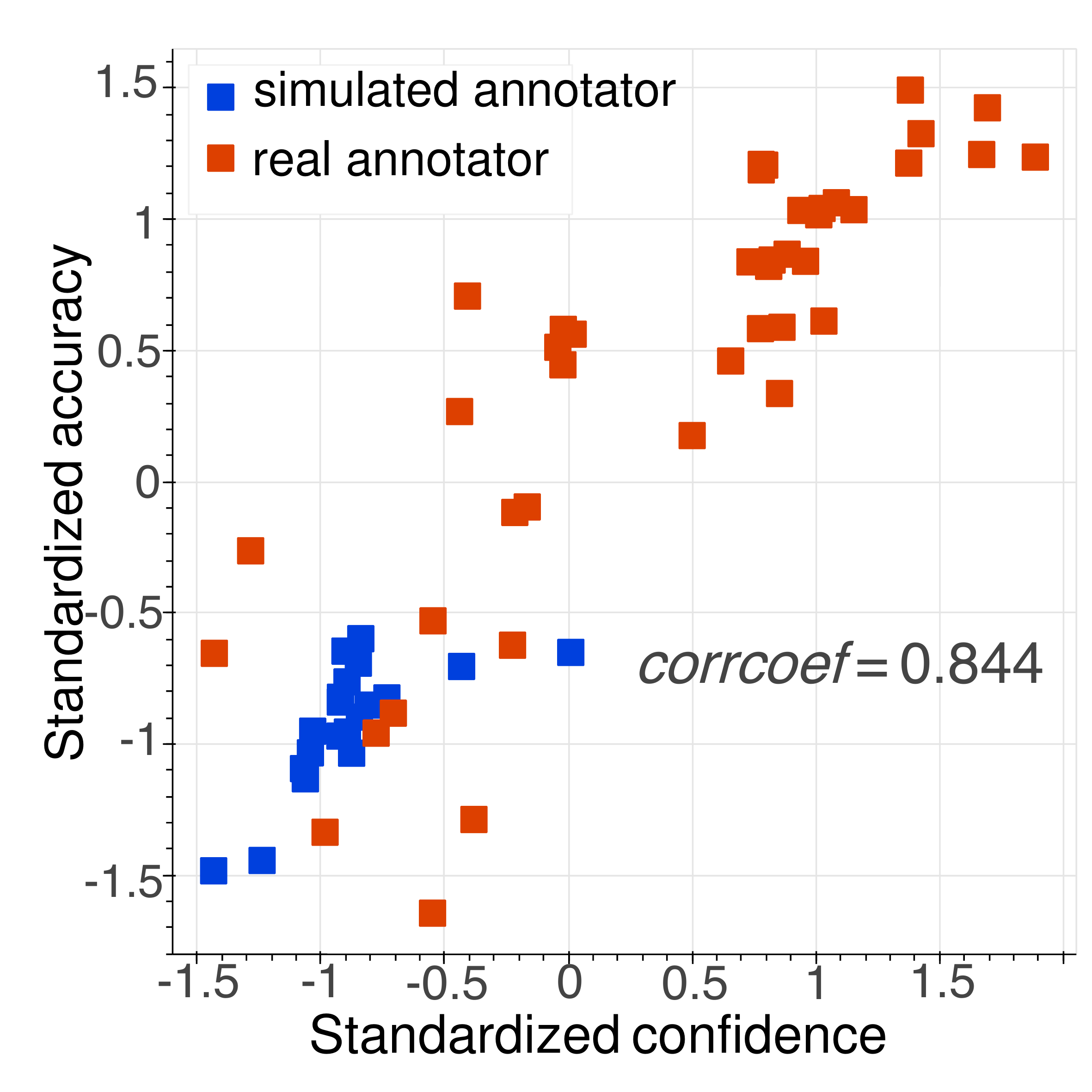}
	\caption{The relations between standardized temporal-aware memory confidence and standardized prediction accuracy of annotators in both real and synthetic data sets. Most of the irresponsible annotators appear in the lower left corner, indicating that there are internal conflicts in their judgments (low confidence), and therefore LR models trained with these labels perform worse than average.}
	\label{fig:acc_confidence}
\end{figure}

We perform standardization on both accuracy scores and the averaged temporal-aware confidence scores within the corresponding data set, and put the standardized values of all the 62 workers from 4 real-world data sets and 4 synthetic data sets together in Figure \ref{fig:acc_confidence}, to reveal the universal relation between temporal-aware confidence and the worker's expertise. We can find a wide range of intra-observer variability among different workers, estimated by their temporal-aware confidence scores. A strong positive correlation is found between averaged confidence and prediction accuracy ($pearson\ r=0.844$). Specifically, synthetic irresponsible annotators, colored in blue, are automatically clustered in the lower left corner, indicating that the poor performances of the classifiers trained with their labels derive from huge inner inconsistencies in their judgments.

\section{Conclusion}
\label{sec:conclusion}
We presented TACMA, an end-to-end framework for language representation learning from crowdsourced labels. Comparing with traditional SRL approaches, the advantages of our framework are: (1) it is able to consider temporal labeling effects within sequences of sample-level labeling tasks for each worker; (2) it automatically computes and aggregates sample-level confidence scores from multiple workers which makes the training process more effective. Experimental results on both synthetic and real-world data sets demonstrates that our approach outperforms other state-of-the-art baselines in terms of accuracy and AUC scores.

\bibliographystyle{acl_natbib}
\bibliography{acl2021}


\end{document}